\documentclass[a4paper, 11pt]{article}

\usepackage[margin=1in]{geometry}
\usepackage{fullpage}

\usepackage{amsmath} \allowdisplaybreaks
\usepackage{amssymb}
\usepackage{mathtools}

\usepackage{natbib}
\usepackage{bibentry}

\usepackage{authblk}

\usepackage{algorithm}
\usepackage{algpseudocode}
\usepackage{multirow}
\usepackage{booktabs} 
\usepackage{mathrsfs}
\usepackage{xcolor}
\usepackage{bbm}
\usepackage{subcaption}
\usepackage{comment}

\usepackage[hidelinks]{hyperref}

\usepackage{tikz}
\usetikzlibrary{positioning,arrows.meta,calc}
\usetikzlibrary{decorations.pathreplacing,arrows.meta, shapes.geometric}
\definecolor{ServiceBlue}{RGB}{68,114,196}
\definecolor{VehicleGreen}{RGB}{16,150,24}
\definecolor{BoxText}{RGB}{255,255,255}
\definecolor{TitleGray}{RGB}{50,50,50}

\definecolor{TD1}{RGB}{ 98,174,87}  
\definecolor{TD2}{RGB}{245,140,75} 

\tikzset{
  veh/.style={
    draw,
    rounded corners=2pt,
    minimum width=1.6cm,
    minimum height=1cm,
    text=white,
    font=\sffamily,
    align=center
  }
}

\usepackage{etoolbox} 
\makeatletter
\def\do#1{\@namedef{#1c}{\ensuremath{\mathcal{#1}}}}
\docsvlist{A,B,C,D,E,F,G,H,I,J,K,L,M,N,O,P,Q,R,S,T,U,V,W,X,Y,Z}
\makeatother
\usepackage{colortbl}
\newcommand{\RNum}[1]{\uppercase\expandafter{\romannumeral #1\relax}}

\author[a]{Abhay Sobhanan}
\author[b]{Hadi Charkhgard}
\author[c]{Changhyun Kwon}
    \affil[a]{Indian Institute of Management Bangalore, Bannerghatta Road, Bengaluru, 560076, India}
	\affil[b]{Department of Industrial and Management Systems Engineering, University of South Florida, Tampa, FL, USA}
    \affil[c]{Department of Industrial and Systems Engineering, KAIST, Daejeon, 34141, Republic of Korea
    }

\title{Arc Routing Problems with Multiple Trucks and Drones: A Hybrid Genetic Algorithm} %

\date{}

\begin{document}

\maketitle

\begin{abstract}
Arc-routing problems underpin numerous critical field operations, including power-line inspection, urban police patrolling, and traffic monitoring. In this domain, the Rural Postman Problem (RPP) is a fundamental variant in which a prescribed subset of edges or arcs in a network must be traversed.
This paper investigates a generalized form of the RPP, called RPP-mTD, which involves a fleet of multiple trucks, each carrying multiple drones. The trucks act as mobile depots traversing a road network, from which drones are launched to execute simultaneous service, with the objective of minimizing the overall makespan.
Given the combinatorial complexity of RPP-mTD, we propose a Hybrid Genetic Algorithm (HGA) that combines population-based exploration with targeted neighborhood searches. Solutions are encoded using a two-layer chromosome that represents: (i) an ordered, directed sequence of required edges, and (ii) their assignment to vehicles. A tailored segment-preserving crossover operator is introduced, along with multiple local search techniques to intensify the optimization.
We benchmark the proposed HGA against established single truck-and-drone instances, demonstrating competitive performance. Additionally, we conduct extensive evaluations on new, larger-scale instances to demonstrate scalability. Our findings highlight the operational benefits of closely integrated truck-drone fleets, affirming the HGA’s practical effectiveness as a decision-support tool in advanced mixed-fleet logistics.
\end{abstract}



\section{Introduction}

Arc Routing Problems (ARPs) represent a fundamental class of optimization problems in logistics, where the objective is to determine optimal routes that traverse specific arcs or edges in a network to satisfy predefined service requirements. In contrast to node routing problems, which focus on visiting designated network nodes, ARPs require servicing along the network's edges or arcs. These problems arise in a wide range of real-world applications, including waste collection, postal delivery, road maintenance, snow plowing, and inspection of electric power transmission lines.
Although node routing problems, such as the Traveling Salesman Problem (TSP) and the Vehicle Routing Problem (VRP), are well-established as $\mathcal{NP}$-hard, ARPs introduce additional complexities, making them at least equally challenging and computationally demanding. As a result, they continue to be a topic of significant interest in combinatorial optimization and logistics research.

In this paper, we focus on a fundamental variant of the ARP known as the Rural Postman Problem (RPP). In the RPP, only a designated subset of edges in the network requires service, while the remaining ``unrequired'' edges are traversed solely to enable more efficient routing. The RPP generalizes the classical Chinese Postman Problem (CPP), in which all edges must be serviced. This added flexibility makes the RPP particularly well-suited to applications such as urban road patrolling, traffic surveillance, and monitoring of energy transmission infrastructure, where selectively traversing unrequired arcs to reach required ones can significantly reduce overall operational costs.

Recent advancements in drone technology \citep{wired2017upsdrone} have given rise to innovative hybrid logistics systems that integrate traditional ground vehicles, such as trucks, with Unmanned Aerial Vehicles (UAVs or drones). This mixed fleet approach significantly enhances operational efficiency \citep{bouman2018dynamic,murray2015flying}, particularly in environments where traditional ground-based logistics are impeded by obstacles or traffic congestion. While drones offer advantages in terms of speed and the ability to bypass ground obstacles, their limited battery capacity inherently restricts their operational range. Integrating drones with trucks addresses this limitation effectively, as trucks can simultaneously serve both as mobile depots, facilitating battery swaps or recharges, and as service vehicles themselves. 

The integration of multiple trucks and drones, each autonomously deploying from and returning to its assigned truck, introduces additional complexities and optimization challenges. Although recent literature has extensively explored such hybrid vehicle-drone systems, existing studies predominantly address node routing problems. Research addressing truck-drone coordination from an arc-routing perspective remains sparse, primarily due to the added computational complexity. One notable exception is provided by \citet{liu2025adaptive}, which investigates a constrained variant of the problem featuring a single truck equipped with drones, serving as a foundation for our broader, multi-vehicle scenario.

Our study contributes to the literature by addressing a novel extension of the traditional RPP involving multiple trucks, each capable of carrying and deploying multiple drones. 
In this setting, each drone must launch from and return to the same truck, creating complex interdependency between truck routes and drone assignments. The primary objective is to minimize the operational makespan, i.e., the total time required to complete all designated required edge services and return all vehicles to the depot, through coordinated truck-drone routing. Furthermore, our approach incorporates realistic operational constraints, including drone flying time limitations imposed by the battery capacity.

Recognizing the inherent computational challenges posed by this problem, we propose a Hybrid Genetic Algorithm (HGA) specifically designed to optimize mixed-fleet coordinated routing for minimal makespan. Our HGA features a distinctive two-part chromosome representation that explicitly encodes the sequencing and directional traversal of required edges, along with their assignments to trucks and drones. Additionally, we enhance the algorithm with multiple effective local search heuristics to systematically guide the exploration of the solution space toward improved optimality. We validate the proposed method through computational experiments on benchmark instances and comparison with an existing algorithm.


The remainder of this paper is organized as follows. Section \ref{sec:lit_review} reviews the related literature, beginning with arc routing problems and followed by drone-assisted routing in both node and arc routing contexts. Section \ref{sec:problem} formally defines the problem, detailing the key assumptions and features. Section \ref{sec:method} describes the proposed hybrid genetic algorithm based solution approach. Computational experiments and performance evaluations are presented in Section \ref{sec:results}. Finally, Section \ref{sec:conclusions} summarizes the key contributions and outlines directions for future research.

\section{Literature Review} \label{sec:lit_review}

In this section, we provide a comprehensive review of the existing literature relevant to our study. 
We begin by examining key research on arc routing problems, highlighting significant contributions in this area. 
Subsequently, we discuss the node routing literature that focuses on the integration of truck-drone mixed fleets. 
Finally, we explore the application of mixed truck-drone fleets within the context of arc routing problems, identifying the existing research gaps that our study aims to address.

\subsection{Arc Routing Problems} 

The family of arc routing problems, arising in various logistics applications \citep{corberan2015arc}, is a crucial area within the routing literature.
For an in-depth review of arc routing problems and their variants, readers may refer to \citet{corberan2021arc}. 
In this study, we focus on the RPP, a fundamental variant of arc routing problems that seeks the shortest tour for a postman to service a designated subset of edges in a network while returning to the starting point.
Notably, when all edges must be traversed, this becomes the CPP, analogous to the TSP in node routing literature.
Several exact methods have been developed for RPP, including \citet{christofides1986algorithm} and \citet{corberan1994polyhedral}, as well as for its variants such as hierarchical \citep{colombi2017hierarchical}, profitable \citep{avila2016branch,colombi2014new}, min-max $k$-vehicle windy \citep{benavent2014branch}, multi-depot \citep{fernandez2018branch} and periodic \citep{benavent2019periodic} RPPs.

However, RPP is $\mathcal{NP}$-hard \citep{lenstra1976general}, necessitating development of near-optimal computational approaches for various practical applications.
Numerous heuristics have been proposed for arc routing problems, such as tabu search \citep{hertz2000tabu}, adaptive large neighborhood search \citep{monroy2017adaptive}, ant colony optimization \citep{santos2010improved}, and genetic algorithms \citep{arakaki2018hybrid,lacomme2001genetic}. 
Genetic algorithms, in particular, have gained popularity as one of the most efficient metaheuristics for routing problems.
\citet{vidal2017node} explore extended neighborhoods and solve the CARP using two approaches: iterated local search and hybrid genetic search. 
A recent study, \citet{mahmoudinazlousolving} propose a genetic algorithm for solving arc routing problems, leveraging dynamic programming decisions to assign arcs to vehicles.

\subsection{Truck-Drone Routing Problems}
We classify the collaborative truck-and-drone routing problems found in the literature into two main categories: node routing and arc routing.

\subsubsection{Node Routing}
The increasing prevalence of drone-assisted deliveries and surveillance has fueled research interest in vehicle routing with drones. 
\citet{murray2015flying} introduced truck routing with drone assistance as the Flying Sidekick Traveling Salesman Problem (FSTSP), presenting two mixed-integer linear programming (MILP) formulations and heuristics. 
A related variant, the Traveling Salesman Problem with Drone (TSP-D), emerged subsequently in \citet{agatz2018optimization}.
Multiple solution methods, including both exact \citep{roberti2021exact} and heuristic \citep{bogyrbayeva2023deep, el2023variable,lee2025iterative,mahmoudinazlou2024hybrid} approaches, have been proposed for the TSP-D.
More generalized extensions of the TSP-D, which incorporate multiple vehicles, have gained attention due to their potential to significantly reduce the makespan.
\citet{tamke2021branch} solve the vehicle routing problem with drones, consisting of multiple truck-drone pairs, using a branch-and-cut method. They solve instances with up to 30 nodes to optimality.
In a recent study, \citet{sobhanan2024branch} address a humanitarian routing problem involving a truck and multiple drones, introducing an exact branch-and-price algorithm tailored to maximize the coverage in emergency logistics. Their results demonstrate the effectiveness of dynamic programming with dominance rules in solving small-sized instances, while also highlighting the scalability limitations of exact optimization approaches for such complex truck-drone coordination problems.
To address larger and more practical instances, heuristic and metaheuristic approaches play a crucial role.
\citet{sacramento2019adaptive} propose an adaptive large neighborhood search method for the vehicle routing problem with multiple trucks, each equipped with one drone.
Similarly, \citet{euchi2021hybrid} present a hybrid genetic algorithm for this problem.
In this work, we focus on an arc routing equivalent of this vehicle routing problem with drones, involving a fleet of multiple trucks and drones.
For comprehensive reviews of collaborative truck and drone routing, one can refer to \citet{chung2020optimization} and \citet{macrina2020drone}. 

\subsubsection{Arc Routing}
\citet{campbell2018drone} addresses the drone arc routing problem, where a drone, unlike traditional vehicles, can travel directly between any two points in the network.
This inherent flexibility renders arc routing with drones a continuous optimization problem. To tackle this, \citet{campbell2018drone} discretizes the problem by representing each edge as a polygonal chain. 
A similar approach is extended for the length-constrained $k$-drones RPP in \citet{campbell2021solving}. 
We incorporate this flexibility of drone traversal into our study by utilizing a distinct graph where edges are constructed based on Euclidean distance.
While the drone arc routing problem has garnered attention, studies on the collaborative truck-drone arc routing problem remain limited. 
\citet{liu2025adaptive} employs a joint system of one truck and one drone to solve the rural postman problem.
The author proposes a formulation for the node routing equivalent of the problem and provide tabu search (TS) and adaptive large neighborhood search (ALNS) algorithms to solve large-scale instances.
Additional heuristic methods have been explored in applications like powerline inspection \citep{liu2019two} and traffic patrolling \citep{luo2019traffic}.

\begin{table}[htbp]
\footnotesize
\centering
\caption{Comparison of related truck-drone routing literature}
\label{tab:lit_rev}
\begin{tabular}{p{4.4cm}lcccccc}
\toprule
\textbf{Reference} & \textbf{Node/Arc} & \textbf{Trucks} & \textbf{Drones} & \textbf{Drone} & \textbf{Method} & \textbf{Problem}\\
\textbf{} & \textbf{Routing} &  & \textbf{} & \textbf{Capacity} & \textbf{} & \textbf{Size}\\
\midrule
\citet{liu2019two} & Arc & 1 & 1 & $D$ & Heuristic & 100 nodes \\ 
\citet{luo2019traffic} & Arc & 1 & 1 & $D$ & Heuristic & 24 nodes \\
\citet{schermer2019matheuristic} & Node & $K$ & $M$ & 1 & Matheuristic & 100 nodes \\
\citet{sacramento2019adaptive} & Node & $K$ & $M$ & 1 & ALNS & 250 nodes \\
\citet{tamke2021branch} & Node & $K$ & $M$ & $D$ & Exact & 30 nodes \\
\citet{euchi2021hybrid} & Node & $K$ & $M$ & 1 & GA & 200 nodes \\
\citet{liu2025adaptive} & Arc & 1 & $2$ & $D$ & TS, ALNS & 542 nodes \\
\citet{mahmoudinazlou2024hybrid} & Node & 1 & 1 & 1 & GA & 250 nodes\\
\midrule
This work &  Arc & $K$ & $M$ & $D$ & GA & 500 nodes \\
\bottomrule
\end{tabular}
\end{table}
 
Table \ref{tab:lit_rev} presents a comparison of existing studies on truck-drone routing problems. Drone capacity $D$ represents the number of nodes or arcs a drone can visit during an independent flight from the truck. To the best of our knowledge, no previous work has proposed a solution approach or solved an arc routing problem involving both multiple trucks and multiple drones. Our study fills this gap by developing a tailored genetic algorithm to solve this problem. Additionally, we extend a traditional assumption of truck-drone routing by allowing drones to traverse multiple arcs, subject to a predefined flying time limit.

\section{Problem Description} \label{sec:problem}

Consider an undirected graph $\Gc = (\Vc, \Ec)$, 
consisting of a set of vertices $\Vc$ and a set of edges $\Ec$ that connect the vertices in $\Vc$. 
Each edge $(i, j) \in \Ec$ is associated with a positive distance or cost, denoted by $\rho(i, j)$.
Furthermore, a subset of edges $\Rc \subseteq \Ec$ is designated as ``required'', meaning these edges must be included in any feasible route, while the remaining edges are optional.
RPP seeks to find a closed tour in $\Gc$ that traverses every required edge in $\Rc$ at least once, while minimizing the total cost of the route. 
The total time is the sum of the travel time of all traversed edges, including any edges that are traversed multiple times.
Since the RPP is known to be a $\mathcal{NP}$-hard problem, finding an optimal solution efficiently is computationally challenging for large instances. 

Numerous algorithms \citep{corberan2000heuristics,monroy2014rural,pearn1995algorithms} have been devised to address the RPP, motivated by its practical significance in various routing contexts.
This study extends the classical RPP to include multiple drone assistance for ground vehicles, a variant we refer to as the RPP with multiple Trucks and multiple Drones (RPP-mTD). 
In this model, the truck acts as a mobile station for each drone designated to it when the drones are not in flight. 
This variant of RPP enhances operational efficiency through coordinated truck-drone operations.
We assume a depot serves as both the starting and ending point for the truck-drone fleet. 
While incorporating a depot node technically transforms the arc routing problem into a general routing problem, we treat it as a special case given the sole involvement of the depot.

The objective function of RPP-mTD is to minimize the makespan, which represents the total time required to complete all services and return the fleet to the depot, while determining the optimal cooperative routes for the truck and drones. 
Every required edge in the network must be serviced at least once by either a truck or a drone. Here, multiple traversals of edges are permitted and is accounted for in the makespan.

\subsection{Key Assumptions}
We explore the RPP-mTD under the following additional assumptions:
\begin{enumerate}
    \item \textit{Fleet Composition:} The fleet consists of one or more trucks ($K$), where each truck carries a fixed number ($M$) of identical drones. The drones are homogeneous in terms of battery capacity and speed, which together determine the maximum allowable flying time for each drone. 
    \item \textit{Truck–Drone Operational Rules:}  A drone may take off from or land on its associated truck only at vertices where the truck is stationed at that moment. Take-off and landing times are assumed to be negligible. In this setup, trucks function as mobile launch and recovery platforms. While drones operate independently during flight, trucks provide essential logistical support, such as instantaneous battery replacement or recharging upon drone landing. Furthermore, when trucks traverse a required edge, they directly service it. 
    \item \textit{Drone Routing Flexibility:} Unlike the trucks, drones are capable of flying directly between vertices without following the predefined road network. A drone can thus bypass ground obstacles by directly flying between any two vertices. To accommodate drone movement, an auxiliary drone-specific graph is implicitly considered, constructed by establishing Euclidean direct edges between all vertex pairs, regardless of road connectivity. When traversing non-required edges, the drone is assumed to follow the corresponding arc in this auxiliary graph. However, when servicing a required arc, the drone instead follows the \emph{shadow} of that arc in the original graph to ensure proper coverage.
    \item \textit{Flight Range Constraint:} Drone flights are restricted by their maximum flight time, limited by battery capacity. The flight duration of any drone sortie cannot exceed this predefined limit. Upon returning to a truck, a drone’s battery is instantly replaced or recharged, resetting its available flight time for subsequent operations. 
    \item \textit{Drone Multi-Arc Visit Flexibility:} Unlike in some known routing problems involving drone assistance, such as the TSP-D, each drone in the RPP-mTD may service multiple required edges consecutively during a single flight, subject to its flight range constraint.
 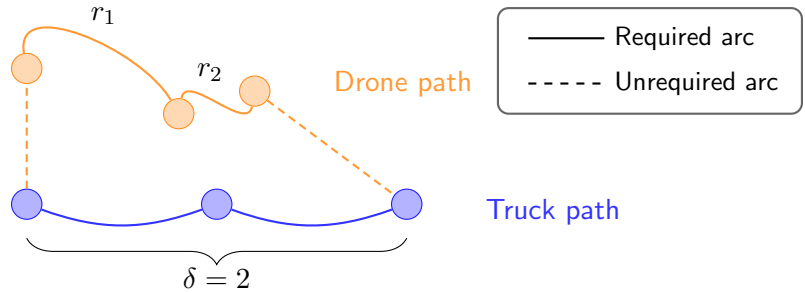
\begin{figure}[htbp]
  \centering
  \begin{tikzpicture}[scale=1.0,
      truckNode/.style={
        circle, draw=blue!90, fill=blue!30, text=white,
        minimum size=4mm, font=\sffamily\bfseries
      },
      droneNode/.style={
        circle, draw=orange!90, fill=orange!30,
        minimum size=4mm
      },
      truckArc/.style={
        thick, draw=blue!80
      },
      droneArc/.style={
        thick, draw=orange!80, densely dashed
      },
      solidDrone/.style={
        thick, draw=orange!80
      },
            droneLbl/.style={font=\scriptsize\sffamily\bfseries, text=orange!80!black},
      lbl/.style={font=\small\sffamily},
      lab/.style={font=\sffamily, inner sep=1pt}
    ]

    \node[truckNode]   (u) at (0,0)   {};
    \node[truckNode]   (v) at (2.5,0) {};
    \node[truckNode]   (w) at (5,0)   {};

    \draw[truckArc]
      (u) to[out=-20,in=-160] (v)
          to[out=-20,in=-160] (w);

    \coordinate (d1) at ($(u)+(0,1.8)$);
    \coordinate (d2) at ($(u)!0.4!(w)+(0,1.2)$);
    \coordinate (d3) at ($(u)!0.6!(w)+(0,1.5)$);

    \node[droneNode] (D1) at (d1) {};
    \node[droneNode] (D2) at (d2) {};
    \node[droneNode] (D3) at (d3) {};

    \draw[droneArc] (u) to (D1);
    \draw[solidDrone]
      (D1) to[out=100,in=120] (D2)
            to[out=80, in=-100] (D3);
    \draw[droneArc] (D3) to (w);

    \node[lab, text=orange!80, anchor=west] at ($(d3)+(1.0,0.1)$)
      {Drone path};
    \node[lab, text=blue!80, anchor=west] at ($(w)+(1.0,-0.1)$)
      {Truck path};
    \node[lab, anchor=west] at ($(d1)+(0.8,0.7)$)
      {$r_1$};
    \node[lab, anchor=west] at ($(d2)+(0.2,0.55)$)
      {$r_2$};

    \draw[decorate, decoration={brace, amplitude=10pt, mirror}]
      ($(u)+(0,-0.45)$) --
      ($(w)+(0,-0.45)$)
      node[midway, below=10pt, lab] {$\delta=2$};
    \begin{scope}[shift={(3.6,4.1)}]
       \draw[rounded corners, thick, black!60] (2.6,-1.5) rectangle (6.6,-2.9);
       \draw[thick, black] (3,-1.9) -- (4,-1.9);
      \node[lbl, anchor=west] at (4,-1.9) {Required arc};
      \draw[thick, black, dashed] (3,-2.5) -- (4,-2.5);
      \node[lbl, anchor=west] at (4,-2.5) {Unrequired arc};
    \end{scope}    
  \end{tikzpicture}
  \caption{An example of the $\delta$‐hop rendezvous mechanism with $\delta = 2$. Here, the drone lands on the last permissible node within the allowed limit.}
  \label{fig:drone_delta}\vspace{-0.5em}
\end{figure}   
    \item \textit{Drone–Truck Coordination Flexibility:} We introduce a hyperparameter, $\delta$, to control the extent of independence between a drone’s movement and its originating truck. When $\delta = 1$, the model adheres to the restricted assumption: the drone is launched and retrieved at consecutive truck nodes, with the truck traversing only one arc during the drone’s flight. In contrast, when $\delta = L$, the truck is permitted to travel up to $L$ arcs after launching the drone before rendezvousing for retrieval. 
Figure \ref{fig:drone_delta} illustrates a scenario where $\delta = 2$ involving one truck and one drone. In this example, the drone returns to the truck at the farthest node permitted within the $\delta$-hop limit.
Increasing $\delta$ significantly enlarges the solution space, as the drone can potentially land at any node visited by the truck within the allowed $\delta$-hop window. This added flexibility can lead to more efficient routing, leading to a reduction in overall makespan. However, practitioners should note that very large values of $\delta$ may introduce practical challenges, such as increased difficulty in real-time communication and a higher risk of coordination failures.
\end{enumerate}

\subsection{An illustrative Example}

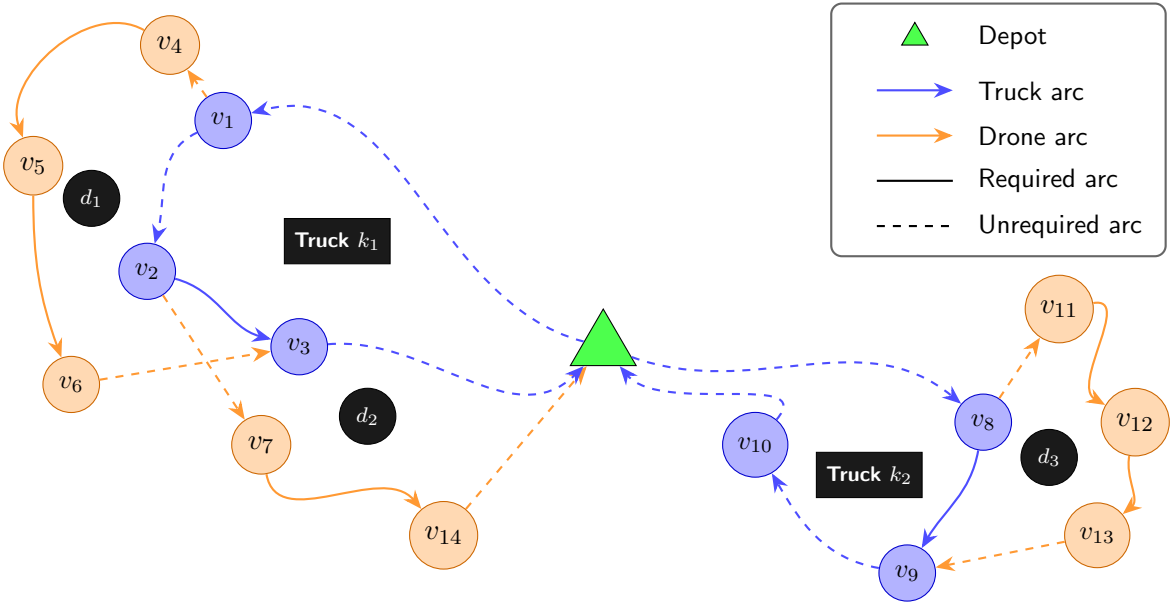
\begin{figure}[htbp]
  \centering
  \begin{tikzpicture}[
      depot/.style={
        regular polygon, regular polygon sides=3,
        minimum size=1cm, draw, fill=green!70, inner sep=0, rotate=0
      },
      truck/.style={
        rectangle, draw=black, fill=black!90,
        minimum width=1cm, minimum height=0.6cm,
        font=\scriptsize\sffamily\bfseries\color{white}
      },
    drone_circle/.style={
        circle, draw=black, fill=black!90,
        minimum width=0.4cm, minimum height=0.4cm,
        font=\scriptsize\sffamily\bfseries\color{white}
      },
      cust1/.style={
        circle, draw=blue!80!black, fill=blue!30,
        minimum size=6mm, font=\small\sffamily
      },
      cust2/.style={
        circle, draw=orange!80!black, fill=orange!30,
        minimum size=6mm, font=\small\sffamily
      },
      drone/.style={
        circle, draw=orange!80!black, fill=orange!30,
        minimum size=6mm
      },
      arcTruck/.style={
        thick, draw=blue!70,
        -{Stealth[length=2.5mm,width=2mm]}
      },
      arcTruckDashed/.style={
        thick, draw=blue!70, dashed,
        -{Stealth[length=2.5mm,width=2mm]}
      },
      arcDrone/.style={
        thick, draw=orange!80, dashed,
        -{Stealth[length=2.5mm,width=2mm]}
      },
      arcDroneSolid/.style={
        thick, draw=orange!80,
        -{Stealth[length=2.5mm,width=2mm]}
      },
      droneLbl/.style={font=\scriptsize\sffamily\bfseries, text=orange!80!black},
      lbl/.style={font=\small\sffamily}
    ]

    \node[depot] (D0) at (0,0) {};

    \node[truck] (K1) at (-3.5,1.4) {Truck $k_1$};
    \node[cust1] (A)  at (-5,3) {$v_1$};
    \node[cust1] (B)  at (-6,1) {$v_2$};
    \node[cust1] (C)  at (-4,0) {$v_3$};

    \node[drone] (d1a) at ($(A)+( -0.7, 1.0)$) {$v_4$};
    \node[drone] (d1b) at ($(B)+(-1.5,1.4)$) {$v_5$};
    \node[cust2] (X)   at ($(B)+(-1.0,-1.5)$) {$v_6$};

    \draw[arcDrone] (A) -- (d1a);
    \draw[arcDroneSolid] (d1a) to[out=150,in=110] (d1b);
    \draw[arcDroneSolid] (d1b) to[out=-90,in=110] (X);
    \draw[arcDrone] (X) -- (C);

    \node[drone_circle,anchor=south west]
          at ($(d1b)+(0.5,-0.7)$) {$d_1$};

    \node[drone] (d2a) at ($(C)+(-0.5,-1.3)$) {$v_7$};
    \node[drone] (d2b) at ($(C)+(1.9,-2.5)$) {$v_{14}$};
    \draw[arcDrone] (B) -- (d2a);
    \draw[arcDroneSolid] (d2a) to[out=-80,in=130] (d2b);
    \draw[arcDrone] (d2b) -- (D0);

    \node[drone_circle,anchor=south]
          at ($(d2a)+(1.4,0.0)$) {$d_2$};

    \draw[arcTruckDashed] (A) to[out=205,in= 75] (B); 
    \draw[arcTruck] (B) to[out=-15,in=165] (C); 
    \draw[arcTruckDashed] (D0) to[out=165,in=15] (A); 
    \draw[arcTruckDashed] (C) to[out=5,in=225] (D0);  

    \node[truck] (K2) at (3.5,-1.7) {Truck $k_2$};
    \node[cust1] (D)  at (5,-1) {$v_8$};
    \node[cust1] (E)  at (4,-3) {$v_9$};
    \node[cust1] (F)  at (2,-1.3) {$v_{10}$};

    \node[drone] (d3) at ($(D)+( 1.0, 1.5)$) {$v_{11}$};
    \node[drone] (d4) at ($(D)+( 2.0, 0.0)$) {$v_{12}$};
    \node[cust2] (Y)  at ($(D)+( 1.5,-1.5)$) {$v_{13}$};

    \draw[arcDrone] (D) -- (d3);
    \draw[arcDroneSolid] (d3) to[out=10,in=150] (d4);
    \draw[arcDroneSolid] (d4) to[out=-100,in=40] (Y);
    \draw[arcDrone] (Y) -- (E);

    \node[drone_circle,anchor=north west]
          at ($(D)+(0.6,-0.2)$) {$d_3$};

    \draw[arcTruckDashed] (D0) to[out=-20,in=140] (D);   
    \draw[arcTruck] (D)  to[out=-100,in=60] (E);          
    \draw[arcTruckDashed] (E) to[out=170,in=-60] (F);     
    \draw[arcTruckDashed] (F) to[out=50,in=-50] (D0);   


    \begin{scope}[shift={(3.6,4.1)}]
      \draw[rounded corners, thick, black!60] (-0.6,0.45) rectangle (3.8,-2.9);
      \def\legendrow#1#2#3{%
        \draw[#1] (0.0,#2) -- (1.0,#2);
        \node[lbl, anchor=west] at (1.2,#2) {#3};
      }
      \node[depot, minimum size=4mm, inner sep=0pt] (legenddepot) at (0.5,0) {};
      \node[lbl, anchor=west] at (1.2,0) {Depot};
      \legendrow{arcTruck}{-0.7}{Truck arc}
      \legendrow{arcDroneSolid}{-1.3}{Drone arc}
      \draw[thick, black] (0.0,-1.9) -- (1.0,-1.9);
      \node[lbl, anchor=west] at (1.2,-1.9) {Required arc};
      \draw[thick, black, dashed] (0.0,-2.5) -- (1.0,-2.5);
      \node[lbl, anchor=west] at (1.2,-2.5) {Unrequired arc};
    \end{scope}

  \end{tikzpicture}
  \caption{An illustrative RPP-mTD solution: two trucks $k_1$ and $k_2$, each carrying two drones—$k_1$ carries $d_1$ and $d_2$; $k_2$ carries $d_3$ and $d_4$. Notably, $d_4$ remains idle in this routing scenario.}
  \label{fig:illustrative_example}
\end{figure}

An illustrative example of a feasible RPP-mTD solution with seven required arcs is shown in Figure~\ref{fig:illustrative_example}. In this scenario, two trucks, $k_1$ and $k_2$, are each equipped with two drones. Specifically, $k_1$ carries drones $d_1$ and $d_2$, while $k_2$ carries drones $d_3$ and $d_4$. Notably, drone $d_4$ remains idle throughout the operation. Among the seven required arcs, two are serviced by drone $d_1$, one by drone $d_2$, one by drone $d_3$, and one by each truck. In this example, the coordination parameter is set to $\delta=2$, allowing a maximum separation of two hops between drone launch and retrieval. Drones $d_1$ and $d_2$ fully utilize this $\delta$-hop flexibility, while drone $d_3$ does not. The figure illustrates the coordinated interaction between trucks and their assigned drones, highlighting how the RPP-mTD framework leverages ground–aerial collaboration to improve efficiency and reduce makespan.

\subsection{Problem Complexity and Formulation Challenges}

The assumptions made in the previous subsection significantly generalize the problem under study compared to the existing body of literature (see arc routing papers in Table~\ref{tab:lit_rev}). This is due to the introduction of several additional flexibility dimensions, including but not limited to the involvement of multiple trucks, multiple drones, drone multi-arc visit flexibility, and drone–truck coordination flexibility.  Unlike classical arc-routing problems or single-truck drone-assisted routing models, RPP-mTD encompasses multiple intertwined decision layers, ranging from vehicle routing and positioning to service assignment and drone endurance constraints. This intricate interplay greatly expands the solution space and makes the problem considerably more challenging.  

We note that developing an integer programming model that fully captures these complexities is highly complicated and provides limited practical value, as such models quickly become intractable for state-of-the-art solvers. Therefore, we do not present an exact optimization model in this paper. Instead, we note that \citet{liu2025adaptive} demonstrate the inherent modeling complexity even for a very special case of our problem (i.e., one truck and one drone), where the computational burden already renders exact methods impractical.   In their study, \citet{liu2025adaptive} transform arc-routing problems into equivalent node-routing formulations \citep{longo2006solving} to address truck–drone coordination, and subsequently design heuristic solution methods due to its computational complexity. While the formulation is effective for their restricted setting, this approach introduces substantial computational overhead by enlarging the underlying graph, which further limits scalability.

Given that our problem setting is far broader, exact optimization techniques are unlikely to be meaningfully applied in real-world scenarios. Additionally, unlike \citet{liu2025adaptive}, we do not rely on node-transformation techniques, which allows us to avoid the additional complexity and overhead associated with graph expansion and an increased fleet size. This observation underscores the need for efficient heuristic approaches capable of producing high-quality solutions within reasonable computational times, which is the main goal of our study.

\section{Methodology} \label{sec:method}

In this section, we present the Hybrid Genetic Algorithm (HGA) developed to solve the RPP-mTD. Our approach builds on established evolutionary computation frameworks while incorporating several novel components that set it apart from standard implementations. These innovations, essential to the algorithm’s effectiveness, are specifically designed to address the unique challenges of coordinated makespan optimization in multi-fleet truck-and-drone routing. The algorithm iteratively applies a combination of traditional and custom genetic operators, complemented by powerful local search procedures, to refine solutions and guide the search toward a global optimum.

\subsection{Graph Abstraction}
We adopt a graph abstraction to enable straightforward solution representation and to appropriately capture truck and drone movements during evaluation. The underlying network consists of both required and unrequired edges. Trucks are constrained to travel along this network topology. Drones, however, operate in a hybrid mode: they fly unconstrained, point-to-point (Euclidean distance) between launch and recovery nodes, but must explicitly traverse the path of any required edge they are assigned to service.

To support the efficient evaluation of several candidate solutions, we preprocess the graph to compute the shortest paths between the endpoints of all required edges. This abstraction reduces the complex pathfinding challenge to a high-level sequencing problem. As a result, the traversal time between any two consecutive service tasks can be retrieved from a precomputed lookup table rather than being repeatedly recomputed, which is critical for the algorithm's performance.

\subsection{Solution Representation}
For an effective optimization process, it is crucial to establish a valid representation of the solution. To capture the distinctive characteristics of the solution space, we adopt a two-part chromosome encoding approach:

\begin{figure}[htbp]
  \centering
  \begin{subfigure}[b]{1.0\linewidth}
    \centering
  \begin{tikzpicture}[
      scale=0.8, transform shape,
      gene/.style={
        draw,
        fill,
        rounded corners=2pt,
        minimum width=1.6cm,
        minimum height=1cm,
        text=BoxText,
        font=\sffamily,
        align=center
      },
      seq/.style={gene, fill=ServiceBlue},
      veh/.style={gene, fill=VehicleGreen},
      arrow/.style={-{Stealth[length=3mm,width=2mm]}, line width=0.8pt}
    ]

    \foreach \i/\val in {1/7,2/1,3/-3,4/5,5/6,6/-4,7/2} {
      \node[seq] (S\i) at ({(\i-1)*2},0) {\val};
    }
    \foreach \i/\val/\col in {
       1/6/TD1,  
       2/2/TD2,  
       3/5/TD1,
       4/3/TD2,
       5/4/TD1,
       6/3/TD2,
       7/6/TD1}{
      \node[veh, fill=\col] (V\i) at ({(\i-1)*2},-1.8) {\val};
    }

    \foreach \i in {1,...,7} {
      \draw[arrow] (V\i.north) -- ++(0,0.3) -| (S\i.south);
    }

    \draw[decorate, decoration={brace, amplitude=10pt}]
      (S1.north west) -- (S7.north east)
      node[midway, yshift=+20pt] {Service Sequence};

    \draw[decorate, decoration={brace, amplitude=10pt, mirror}]
      (V1.south west) -- (V7.south east)
      node[midway, yshift=-20pt] {Vehicle Assignments};
  \end{tikzpicture}
  \caption{Two‐part chromosome encoding: the service sequence of required edges (top) and vehicle assignments (bottom).}
  \label{fig:two_part_chromosome}
  \end{subfigure}
  
  \vspace{1em}
  
  \begin{subfigure}[b]{1.0\linewidth}
    \centering
    \begin{tikzpicture}[
        scale=0.75, transform shape,
        gene/.style={
          draw,
          rounded corners=2pt,
          minimum width=1.6cm,
          minimum height=1cm,
          text=black,
          font=\sffamily,
          align=center
        },
        svc2/.style={gene, fill=ServiceBlue, text=BoxText},
        drv2/.style={gene, fill=TD1},
        arrow/.style={-{Stealth[length=3mm,width=2mm]}, line width=0.8pt}
      ]
      \foreach \i/\val in {1/7,2/-3,3/6,4/2} {
        \node[svc2] (T\i) at ({(\i-1)*2},0) {\val};
      }
      \foreach \i/\val in {1/6,2/5,3/4,4/6} {
        \node[drv2] (D\i) at ({(\i-1)*2},-1.8) {\val};      }
      \foreach \i in {1,2,3,4} {
        \draw[arrow] (D\i.north) -- ++(0,0.3) -| (T\i.south);
      }
    \end{tikzpicture}
    \caption{Decomposition for Truck-Drone System 2}
    \label{fig:decomposition}
  \end{subfigure}

  \caption{An example solution representation of an RPP-mTD with $R=7$, $K=2$ and $M=2$}
  \label{fig:chromosome}
\end{figure}
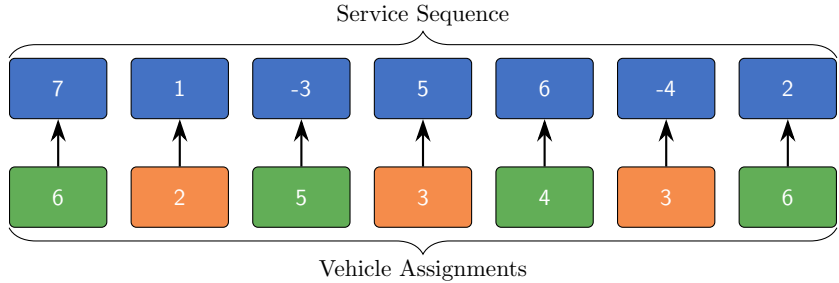
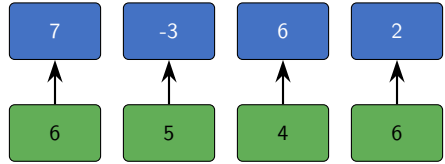

\begin{enumerate}
    \item \textit{Service Sequence:} This component is an ordered list representing all required edges that must be traversed, each uniquely indexed. With $R$ required edges, this chromosome segment has length $R$. Each required edge is accompanied by a direction marker to indicate traversal direction: a positive index denotes forward traversal, while a negative sign indicates traversal in the opposite direction. Collectively, this chromosome part defines the sequence of service and traversal orientations.

    \item \textit{Vehicle Assignment:} This part complements the service-order sequence by specifying the vehicle responsible for each service step. Each vehicle is assigned a unique identifier to facilitate differentiation. For example, in a configuration with $K=2$ trucks, each equipped with $M=2$ drones, the numbering proceeds as follows: Truck 1 is labeled as 1, its drones as 2 and 3, Truck 2 as 4, followed by its drones as 5 and 6, respectively. Figure \ref{fig:chromosome} shows an example of the chromosome encoding for a problem with $R=7$, illustrating how the representation can be decomposed for a truck system. Tasks assigned to a truck and its associated drones collectively constitute the tour of that truck-drone system, and those allocated specifically to individual drones form their respective sorties.
\end{enumerate}

Since both parts of the chromosome are of equal length and jointly capture all necessary information, a complete routing plan for the vehicles can be derived directly by interpreting the chromosome. This representation clearly separates \textit{what} must be done (the exact sequence and orientation of required arcs) from \textit{which} vehicle performs it, while maintaining full synchronization between the two components.

\subsection{Route Construction for Evaluation}
To translate a chromosome into a complete routing plan and evaluate its quality, we follow a clear decoding process.
First, the full route for each vehicle is constructed by connecting its assigned service tasks. When connecting two consecutive tasks, if multiple precomputed shortest paths of equal length (co-optimal paths) exist, one is randomly sampled. This random choice occurs during the creation and modification of solutions, introducing valuable structural diversity into the population by exploring different intermediate nodes for drone operations.

Second, the process handles structural inefficiencies implicitly. In cases where the shortest path between required edges $r_i$ and $r_j$ happens to pass through another required edge $r_k$, the resulting route is valid but contains a redundancy. Such solutions are not explicitly forbidden or repaired. Instead, their naturally longer makespan results in a poorer fitness score, and they are less likely to be selected for propagation, allowing the evolutionary search to organically favor more efficient orderings (e.g.,~...$r_i, r_k, r_j$...). The quality of a decoded plan is ultimately measured by its makespan, which serves as the primary input to the fitness function described in the following section.

\subsection{Fitness Function}

The fitness of each candidate solution is measured by its makespan, i.e., the total time required to service all designated edges and return to the depot. To compute this makespan, we first determine each truck’s route duration by summing the travel times along the shortest paths that connect the pairs of required arcs in its service sequence (including any necessary repeats of edges). Next, we calculate each drone’s flight time using the straight-line distance between non-required edges, based on the potential launch and recovery vertices. For required edges, the drone’s travel time is computed along the given edges. We also verify that every drone route respects the maximum allowable flight time, $\tau$, and does not exceed the permitted number of intervening truck-traversal arcs, $\delta$, between take-off and landing points. 
Finally, the drone paths are selected based on their effect on the objective value.

After the makespan $T(I)$ of an individual $I$ has been calculated, we translate it into a fitness score $F(I)$. 
Note that any solution violating the drone‐range constraint incurs a range‐violation penalty
\[
w_{\mathrm{inf}}
\sum_{d=1}^{D}
\bigl(\tau_{\max}(d) - \tau \bigr)^{+}  
\]
added directly to its makespan.
Here, \(\tau_{\max}(d)\) denotes the highest independent flight time of drone \(d\), and \((x)^{+} = \max\{0, x\}\). 
$w_{\rm inf}$ is the penalty that we dynamically adjust throughout the genetic algorithm.
To encourage population diversity, we also compute each individual’s diversity score $\delta(I)$ as the normalized Hamming distance to its two nearest neighbors \citep{sobhanan2024equity}. 
If $n_E$ denotes the number of elite individuals in the population and $n_P$ is the total population size, then the fitness function for a chromosome $I$ is defined as
\[
F(I) \;=\; T(I)\, \times \!\biggl(\frac{n_E}{n_P}\biggr)^{\delta(I)}.
\]
Therefore, solutions with lower fitness values, reflecting both shorter effective makespans and greater niche differentiation, are preferred in the selection process.

\subsection{Initial Population}

The initial population of solutions is generated using a hybrid strategy that combines purely random initialization with targeted initialization.  Specifically, a fraction \(p_t\) of the population is seeded by a problem‐specific heuristic that identifies promising drone–truck routing plans. By incorporating these makespan‐aware individuals alongside randomly generated solutions, the genetic algorithm benefits from both high‐quality starting points and broad population diversity.

Each targeted individual is constructed in two steps.  First, we build a simple RPP service sequence: starting at the depot, we iteratively append the as‐yet‐unserved required edge whose nearest endpoint to the current truck location that minimizes the incremental truck travel time, while respecting edge directionality by choosing the traversal order \((u,v)\) or \((v,u)\) that is most efficient.  Second, we apply a greedy makespan‐minimization assignment.  We traverse the resulting service sequence and, for each arc \((u,v)\), compute the projected completion time if served by each of the \(K\) trucks or any of their \(K\times M\) drones, tracking each truck’s current node and accumulated drive time, along with each drone’s cumulative flight time.  We then assign the arc to the vehicle that yields the smallest overall makespan.

\subsection{Evolutionary Process}
The population evolves through an iterative evolutionary process involving selection, crossover, and mutation, executed for a predefined maximum number of generations \(G\). At each generation, offspring solutions are generated by combining existing solutions through genetic operations, ensuring exploration of diverse and potentially superior solutions.

Crossover operators play a crucial role by combining chromosome sequences from pairs of parent solutions to produce offspring inheriting traits from both parents. Parent solutions are selected using binary tournament selection based on their fitness scores. We randomly apply one of three crossover methods. The \textit{Order Crossover (OX)} method involves selecting two cut points in the chromosome sequence, copying the segment between these points from one parent into the offspring, and filling the remaining positions sequentially with genes from the other parent, skipping genes already copied. The \textit{Partially Mapped Crossover (PMX)} exchanges segments between two cut points while resolving duplicate genes through a mapping process between parents, ensuring valid offspring permutations.

Additionally, we propose a novel, problem-specific crossover operator termed \textit{Segment-Preserving Crossover}, tailored explicitly for the truck-drone system context. This operator selects a complete segment corresponding to all visits handled by a single truck system and its associated drones from one parent, and extracts the corresponding segment from the other parent. Subsequently, either OX or PMX is applied exclusively to these segments. The generated segment is then reintegrated into copies of the original parents, followed by a repair procedure. The repair step corrects any duplicate or missing visits by incorporating appropriate arcs from the opposite parent. Consequently, this method yields two feasible offspring preserving intact truck system-level segments.

Mutation operations occur with an initial probability \( p_m \), introducing minor random variations within chromosome structures to explore new solution spaces and maintain genetic diversity. To mitigate stagnation, if no improvement in solution quality is observed for \( G_m \) consecutive generations, we dynamically increase the mutation probability to \( p_m^+ \). Mutation is executed by randomly selecting one of three operators applied to both the service sequence and vehicle assignment components of the chromosome: (1) \textit{Swap Mutation} selects two positions at random and exchanges their genes, (2) \textit{Inversion Mutation} selects a random subsequence between indices \( i < j \) and reverses it in place, and (3) \textit{Reassignment Mutation}, designed specifically for vehicle assignments, randomly selects a single gene and assigns it to a different vehicle.

Throughout the evolutionary process, the population size is maintained within the interval \([P_L, P_H]\). After generating offspring solutions at each generation, the population is sorted based on the fitness function, and only the best-performing \( P_L \) individuals are retained for subsequent generations. This strategy effectively balances diversity with exploration of high-quality solutions. Additionally, a maximum of \( P_H \) candidate solutions is generated at each iteration through combined evolutionary operations and local search refinements.
To further promote convergence, the top 1\% of the solutions are explicitly preserved across generations. This elitist strategy ensures that superior solutions are not lost due to stochastic variation.

\subsection{Local Search and Refinement}

To further enhance exploration and mitigate the risk of becoming trapped in local optima, each newly generated population undergoes a dedicated local improvement phase. Following the evaluation of offspring fitness, individuals are sorted in ascending order of their fitness scores, and the top 20\% are selected for further refinement. Each selected individual is subjected to a bounded iterative local search process, limited to a number of steps ($\texttt{ls\_steps}$). This process involves randomized exploration of the solution’s neighborhood, and unlike the mutation procedure, only strictly improving moves are accepted. This ensures a monotonically non-increasing makespan and steady progression toward local optimality.

At each iteration of the local search, a candidate solution is modified using one randomly chosen neighborhood search operator from the following five strategies:

\begin{enumerate}
\item \textit{Subsequence Reversal}: Selects and reverses a subsequence of the overall service sequence and the corresponding vehicle assignments, to potentially reveal more efficient orderings.

\item \textit{Or-opt}: Relocates a contiguous block of up to $b$ required arcs to a different position in the sequence, preserving their internal order, to exploit local structural improvements.

\item \textit{Drone Sortie Optimization}: For each drone, attempts to improve the internal ordering of assigned arcs when the sortie length exceeds a threshold (e.g., $\geq 3$), effectively solving small intra-sortie Traveling Salesman subproblems.

\item \textit{Greedy Vehicle Reassignment}: Evaluates alternative vehicle assignments for an arc, identifies the reassignment yielding the greatest reduction in makespan, and applies the most beneficial changes.

\item \textit{Ruin-and-Construct}: Temporarily removes a proportion $p_\text{ruin}$ of the required arcs and reinserts them into positions and assigns vehicles that minimize the resulting makespan, leading to additional exploration of the solution space.

\end{enumerate}

After applying a neighborhood move, the solution’s fitness is re-evaluated. If the resulting makespan is strictly better than that of the incumbent solution, the candidate replaces it. This process is repeated for up to $\texttt{ls\_steps}$ iterations, after which the locally refined individual is returned.
Once all selected candidates in the current generation have undergone local refinement, the resulting pool is trimmed based on fitness to restore the original population size.

By integrating broad global search via genetic evolution with focused local refinement, our hybrid genetic algorithm (HGA) strikes a robust balance between diversification and exploitation. This balance leads to high-quality solutions with minimized makespan. Upon convergence, the best solution found across all generations is selected as the final output, representing a practically deployable multi-fleet truck-drone routing plan.

\section{Computational Study} \label{sec:results}

In this section, we present a series of experiments to evaluate the performance of the proposed heuristic for the RPP-mTD under various problem settings.
All algorithms and experimental procedures are implemented in Julia 1.11.5 and executed on a MacBook Pro equipped with an Apple M1 chip, 16 GB of RAM, and running macOS Sequoia 15.5. 
The parameters used in our experiments are as follows: \( n_E = 0.8n_P \), \( G = 100 \), \( G_m = 10 \), \( P_L = 100 \), \( P_H = 200 \), \( w_{\text{inf}} \in [0.01, 100.0] \), \( p_t = 0.1 \), \( p_m = 0.1 \), \( p_m^+ = 0.3 \), \( p_{\text{ruin}} = 0.2 \), and $\texttt{ls\_steps} = 30$.

We conduct two types of experiments. First, we benchmark our algorithm using the instances provided in \citet{liu2025adaptive}, focusing on a restricted setting with $K = 1$ truck and $M = 1$ drone. 
This allows for a controlled comparison of a special case with the existing method, using one of their parameter configurations with the drone speed ($s_\text{drone} = 2$) and the truck speed ($s_\text{truck} = 1$), the most common situation in practical applications \citep{harrison2025}. 
Second, we generate a set of large-scale RPP-mTD instances to analyze the algorithm’s scalability and responsiveness to different problem characteristics. Specifically, we examine the effects of varying vehicle compositions, drone ranges, and the $\delta$-hop window. In these experiments, truck travel time on an edge is calculated by dividing its arc length by a constant cruise speed of \(40~\text{km}\,\text{h}^{-1}\).  
Since drones are generally faster than trucks \citep{harrison2025}, especially in urban environments, their flight time is computed using a cruise speed of \(80~\text{km}\,\text{h}^{-1}\) \citep{aurambout2022drone}.
Each sortie is further limited by a maximum airborne time of \(60\) minutes, reflecting the battery capacity constraints.
\color{black}

For the second set of experiments, we generate RPP test instances based on a \(10 \text{ km}\times 10 \text{ km}\) square grid that approximates an urban district.  
First, the number of nodes $N$ is selected such that $N \in [50, 500]$, and node coordinates are generated randomly on the square grid.  
A graph is then constructed by connecting the nodes using random edges.
To ensure that routes exist, we retain the largest connected component and discard any isolated nodes or subgraphs.  
A subset of these edges is then designated as required, with its cardinality selected from the range $[15, 100]$.  
All edge lengths in the ground network are measured using the Manhattan metric to realistically capture rectilinear street layouts.  
For each instance size considered, we generate five test instances per instance class. 

\subsection{Benchmark Evaluation}

In Table~\ref{tab:benchmark_m1}, we present a benchmark evaluation of our proposed approach applied to the special case of the RPP-mTD problem with a single truck carrying one drone. We utilize benchmark instances originally introduced in \citet{liu2025adaptive}. In this experiment, we do not impose a limit on the $\delta$-hop window. Additionally, we analyze the impact of varying values of $\beta$, a hyperparameter from their study, to determine the drone flying time limit, defined as follows:
\color{black}

\begin{equation}
    \tau = \frac{\beta}{s_\text{drone}} \times \text{Average edge distance in the drone graph}
\end{equation}

\begin{table}[htbp]
\scriptsize
\centering
\caption{Benchmark evaluation results for RPP-mTD with 1 truck and 1 drone}
\label{tab:benchmark_m1}
\begin{tabular}{cccrrrrrr}
\toprule
\multirow{2}{*}{Instance Class} & \multicolumn{1}{c}{\multirow{2}{*}{$\beta$}} & \multicolumn{1}{c}{\multirow{2}{*}{Best Known}} & \multicolumn{3}{c}{ALNS}           & \multicolumn{3}{c}{HGA}     \\
\cmidrule(lr){4-6} \cmidrule(lr){7-9}
& \multicolumn{1}{c}{}     & \multicolumn{1}{c}{}           & \multicolumn{1}{c}{Obj} & \multicolumn{1}{c}{Time*} & 
\multicolumn{1}{c}{Gap} & 
\multicolumn{1}{c}{Obj} & \multicolumn{1}{c}{Time}  & \multicolumn{1}{c}{Gap}  \\
\multicolumn{1}{l}{}            &         & \multicolumn{1}{l}{}           & \multicolumn{1}{l}{}    & \multicolumn{1}{c}{(sec)} & \multicolumn{1}{c}{(\%)} & \multicolumn{1}{l}{}    & \multicolumn{1}{c}{(sec)} & \multicolumn{1}{c}{(\%)} \\
\midrule
\multirow{3}{*}{N10E20R5}       & 1& 358.22 & 365.32 & 11.23    & 1.98    & 368.82 & 2.82     & 2.96    \\
        & 2& 256.43 & 263.23 & 7.32     & 2.65    & 260.80 & 3.15     & 1.70    \\
        & 3& 196.29 & 200.42 & 5.43     & 2.10    & 196.76 & 3.15     & 0.24    \\
        \hline
\multirow{3}{*}{N10E20R7}       & 1& 461.53 & 470.34 & 16.23    & 1.91    & 475.62 & 6.92     & 3.05    \\
        & 2& 319.89 & 326.43 & 34.23    & 2.04    & 321.96 & 7.97     & 0.65    \\
        & 3& 237.08 & 242.53 & 43.42    & 2.30    & 242.82 & 8.48     & 2.42    \\
        \hline
\multirow{3}{*}{N10E20R10}      & 1& 591.06 & 599.42 & 284.55   & 1.41    & 629.20 & 16.39    & 6.45    \\
        & 2& 394.15 & 400.23 & 204.32   & 1.54    & 399.86 & 22.12    & 1.45    \\
        & 3& 284.21 & 289.53 & 221.23   & 1.87    & 289.26 & 24.99    & 1.78    \\
        \hline
\multirow{3}{*}{N15E30R5}       & 1& 446.42 & 452.32 & 4.63     & 1.32    & 456.50 & 2.77     & 2.26    \\
        & 2& 340.98 & 344.52 & 7.83     & 1.04    & 349.48 & 2.08     & 2.49    \\
        & 3& 253.60 & 259.32 & 6.58     & 2.26    & 263.64 & 1.94     & 3.96    \\
        \hline
\multirow{3}{*}{N15E30R7}       & 1& 533.58 & 547.63 & 19.43    & 2.63    & 552.46 & 6.44     & 3.54    \\
        & 2& 399.35 & 410.23 & 35.64    & 2.72    & 404.54 & 5.21     & 1.30    \\
        & 3& 282.83 & 289.53 & 55.47    & 2.37    & 287.80 & 4.78     & 1.76    \\
        \hline
\multirow{3}{*}{N15E30R10}      & 1& 660.18 & 675.73 & 210.24   & 2.36    & 690.34 & 16.45    & 4.57    \\
        & 2& 481.08 & 510.23 & 199.53   & 6.06    & 481.08 & 20.73    & 0.00    \\
        & 3& 337.74 & 359.34 & 178.46   & 6.40    & 337.56 & 22.19    & -0.05   \\
\midrule
Average &  & 379.70 & 389.24 & 85.88    & 2.50    & 389.36 & 9.92     & 2.25   \\
\bottomrule
\end{tabular}
\end{table}

The results indicate that our method achieves competitive performance compared to the ALNS. Specifically, our method has an average optimality gap of $2.25\%$, closely matching ALNS, which achieves a gap of $2.5\%$. It is worth emphasizing that while ALNS is specifically tailored for single-truck routing scenarios, our solution method is more general, capable of addressing broader configurations of the problem.
Moreover, our approach demonstrates notable advantages over existing benchmarks. For instance, in one specific scenario (instance class N15E30R10 with $\beta=3$), our method outperforms the best-known solution obtained using a MILP solver that was executed with a one-hour time limit. Additionally, our algorithm significantly reduces computational time compared to ALNS, delivering solutions significantly faster. 
Note, however, that the ALNS results were reported in the benchmark study using a workstation equipped with a 2.2 GHz Intel Xeon E5-2630 processor and 32 GB of RAM. 
Using data from \url{https://www.cpubenchmark.net/}, we compared the single-thread performance of the Xeon processor to that of our system. Based on this comparison, we estimate that if ALNS were executed on our hardware, its average runtime would be approximately 40.3 seconds, as opposed to the value reported in the benchmark study. Nevertheless, even under this adjusted estimate, our HGA algorithm maintains a clear advantage in computational efficiency, delivering solutions in significantly less time.

\subsection{Scalability and Responsiveness Evaluation}

\begin{figure}
    \centering
    \includegraphics[width=0.65\linewidth]{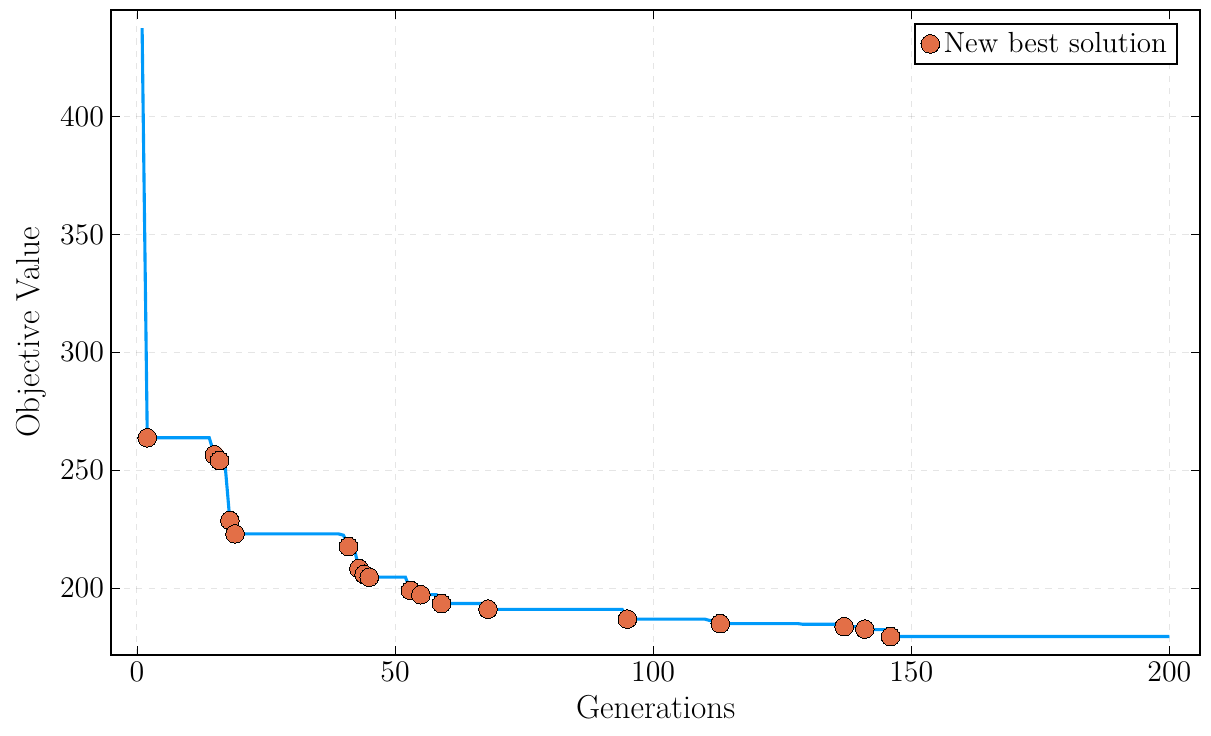}
    \caption{Convergence of the proposed HGA across generations for an instance with $R = 100$}
    \label{fig:convergence_R100}
\end{figure}

We now conduct experiments on larger instances to better understand and empirically evaluate the factors that influence both the RPP-mTD and our proposed HGA. Before presenting the various experiments and results in this subsection, we first highlight the performance of the proposed HGA on a representative instance. This provides evidence that the algorithm does not waste computational resources; rather, it continuously searches for improved solutions until convergence.  Figure~\ref{fig:convergence_R100} illustrates the convergence behavior of the HGA over generations for an instance with $R = 100$, using a configuration of $K = 2$ trucks, $M = 2$ drones per truck, and a drone flying time limit of $\tau = 1.0$ hour. As observed, the proposed method quickly identifies a feasible solution early in the search with an objective value of approximately 265. At approximately 150 generations, it finds the best solution with an objective value near 180, corresponding to an improvement of roughly 32\%. After 200 iterations, the algorithm terminates and reports the best solution found.

\subsubsection{Analysis of the Impact of Vehicle Composition}

In this experiment, we analyze how the makespan of the system changes with respect to different combinations of trucks and drones, in addition to the computational complexity it possesses as we increase the fleet size. 
Tables~\ref{tab:varying_veh_1} and \ref{tab:varying_veh_2} present the results of our method applied to varying vehicle compositions, specifically with $K \in \{1, 2, 3\}$ trucks and $M \in \{1, 2, 3\}$ drones. 
All the experiments were conducted using fixed parameters $\delta = 5$ and $\tau = 1.0$ hour. Using our proposed HGA, we examine how changes in the number of trucks and drones influence solution quality and computational runtime.

\begin{table}[htbp]
\scriptsize
\centering
\caption{Impact of vehicle composition on solution quality and runtime - Part 1}
\label{tab:varying_veh_1}
\resizebox{\textwidth}{!}{
\begin{tabular}{crrrrrrrrrrrr}
\toprule
\multirow{3}{*}{Instance} & \multicolumn{6}{c}{1 Truck}       & \multicolumn{6}{c}{2 Trucks}      \\
\cmidrule(lr){2-7} \cmidrule(lr){8-13}
        & \multicolumn{2}{c}{$M = 1$}         & \multicolumn{2}{c}{$M = 2$}         & \multicolumn{2}{c}{$M = 3$}         & \multicolumn{2}{c}{$M = 1$}         & \multicolumn{2}{c}{$M = 2$}         & \multicolumn{2}{c}{$M = 3$}         \\
        \cmidrule(lr){2-3} \cmidrule(lr){4-5} \cmidrule(lr){6-7} \cmidrule(lr){8-9} \cmidrule(lr){10-11} \cmidrule(lr){12-13}
        & \multicolumn{1}{c}{Obj} & \multicolumn{1}{c}{Time}  & \multicolumn{1}{c}{Obj} & \multicolumn{1}{c}{Time}  & \multicolumn{1}{c}{Obj} & \multicolumn{1}{c}{Time}  & \multicolumn{1}{c}{Obj} & \multicolumn{1}{c}{Time}  & \multicolumn{1}{c}{Obj} & \multicolumn{1}{c}{Time}  & \multicolumn{1}{c}{Obj} & \multicolumn{1}{c}{Time}  \\
\multicolumn{1}{l}{}      & \multicolumn{1}{l}{}    & \multicolumn{1}{l}{(min)} & \multicolumn{1}{l}{}    & \multicolumn{1}{l}{(min)} & \multicolumn{1}{l}{}    & \multicolumn{1}{l}{(min)} & \multicolumn{1}{l}{}    & \multicolumn{1}{l}{(min)} & \multicolumn{1}{l}{}    & \multicolumn{1}{l}{(min)} & \multicolumn{1}{l}{}    & \multicolumn{1}{l}{(min)} \\
\midrule
N50E100R15 & 33.8   & 0.2      & 32.2   & 0.3      & 32.2   & 0.3      & 20.8   & 0.3      & 20.5   & 0.4      & 20.4   & 0.5      \\
N50E100R20 & 45.4   & 0.4      & 44.3   & 0.5      & 42.2   & 0.6      & 27.0   & 0.6      & 26.8   & 0.8      & 25.2   & 1.0      \\
N100E200R25& 70.1   & 0.6      & 68.4   & 0.9      & 65.7   & 1.0      & 42.2   & 1.1      & 38.9   & 1.2      & 38.9   & 1.6      \\
N100E200R30& 80.2   & 0.9      & 77.1   & 1.4      & 76.9   & 1.7      & 48.4   & 1.8      & 45.1   & 2.3      & 44.4   & 3.0      \\
N200E400R40& 122.2  & 2.1      & 112.1  & 3.3      & 109.9  & 4.0      & 67.9   & 4.1      & 63.3   & 5.0      & 61.9   & 6.5      \\
N200E400R50& 139.2  & 4.2      & 136.1  & 6.3      & 130.5  & 7.6      & 81.6   & 7.6      & 78.0   & 9.4      & 74.7   & 11.9     \\
N300E600R50& 148.6  & 4.4      & 142.2  & 6.7      & 139.6  & 8.1      & 85.8   & 8.0      & 83.0   & 10.3     & 81.3   & 12.6     \\
N300E700R70& 203.7  & 11.8     & 189.4  & 17.6     & 188.2  & 20.8     & 112.8  & 20.7     & 108.0  & 25.7     & 104.3  & 32.0     \\
N400E800R80& 267.7  & 18.4     & 251.8  & 27.5     & 242.5  & 32.0     & 149.8  & 32.8     & 143.5  & 40.1     & 135.8  & 49.8     \\
N500E1000R100             & 321.0  & 38.5     & 312.8  & 57.2     & 302.6  & 65.8     & 186.0  & 66.7     & 174.5  & 83.2     & 168.3  & 98.6   \\
\bottomrule
\end{tabular}
}
\end{table}

\begin{table}[htbp]
\centering
\scriptsize
\caption{Impact of vehicle composition on solution quality and runtime - Part 2}
\label{tab:varying_veh_2}
\begin{tabular}{crrrrrr}
\toprule
\multirow{3}{*}{Instance} & \multicolumn{6}{c}{3 Trucks} \\
\cmidrule(lr){2-7}
       & \multicolumn{2}{c}{$M = 1$}        & \multicolumn{2}{c}{$M = 2$}        & \multicolumn{2}{c}{$M = 3$}        \\
       \cmidrule(lr){2-3} \cmidrule(lr){4-5} \cmidrule(lr){6-7}
       & \multicolumn{1}{c}{Obj} & \multicolumn{1}{c}{Time}  & \multicolumn{1}{c}{Obj} & \multicolumn{1}{c}{Time}  & \multicolumn{1}{c}{Obj} & \multicolumn{1}{c}{Time}  \\
\multicolumn{1}{l}{}      & \multicolumn{1}{l}{}    & \multicolumn{1}{l}{(min)} & \multicolumn{1}{l}{}    & \multicolumn{1}{l}{(min)} & \multicolumn{1}{l}{}    & \multicolumn{1}{l}{(min)} \\
\midrule
N50E100R15 & 16.4   & 0.5      & 16.3   & 0.6      & 16.1   & 0.6      \\
N50E100R20 & 20.7   & 0.9      & 20.2   & 1.0      & 19.7   & 1.3      \\
N100E200R25& 31.7   & 1.7      & 29.8   & 1.9      & 29.4   & 2.3      \\
N100E200R30& 35.4   & 2.8      & 34.4   & 3.2      & 34.1   & 3.8      \\
N200E400R40& 48.5   & 6.3      & 46.2   & 7.1      & 46.0   & 8.8      \\
N200E400R50& 59.0   & 11.9     & 56.0   & 13.4     & 55.5   & 16.3     \\
N300E600R50& 64.6   & 12.3     & 58.9   & 14.1     & 58.9   & 17.2     \\
N300E700R70& 82.4   & 32.8     & 76.1   & 36.1     & 76.1   & 43.8     \\
N400E800R80& 109.5  & 49.1     & 102.2  & 55.2     & 100.5  & 65.3     \\
N500E1000R100             & 131.2  & 97.8     & 125.2  & 112.0    & 121.2  & 131.3   \\
\bottomrule
\end{tabular}
\end{table}

Across all instance classes, increasing the number of trucks consistently led to substantial reductions in the makespan. For example, in instance N200E400R50, transitioning from a single truck with one drone to a three-truck setup, each still equipped with one drone, reduced the makespan by approximately 57.6\%, from 139.2 to 59.0. Similar patterns were observed across all problem sizes, underscoring the advantage of using multiple trucks for parallel coverage, particularly in larger networks.
Adding drones per truck generally improved solution quality as well, although the marginal gains diminished with each additional drone due to drone range limitations. Introducing a second drone per truck consistently yielded noticeable improvements in the objective value. However, the benefit of adding a third drone was often limited, especially in larger, multi-truck configurations. For instance, in the N400E800R80 case with three trucks, the improvement in the objective was only 1.7\%, when increasing from two to three drones per truck, despite a significant increase in computational runtime from 55.2 to 65.3 minutes.

From a computational perspective, the runtime of our HGA algorithm increases significantly with the number of trucks and drones. This overhead becomes especially noticeable as more drones or trucks are added, reflecting the increased complexity of coordinating multiple drone routes with corresponding truck schedules. This effect is especially evident in larger instances. In the N500E1000R100 scenario, for example, the runtime nearly doubled, from 57.2 to 112.0 minutes, when moving from a single-truck, two-drone setup to a three-truck configuration with two drones each.
These empirical results clearly highlight the trade-offs between solution quality and computational complexity. While deploying multiple trucks consistently enhances performance by enabling parallel operations, drone allocation requires greater caution due to their limited flight range. Given the diminishing marginal gains and the sharp increase in computational burden, a balanced strategy is essential when configuring larger fleets.

\subsubsection{Analysis of Drone Flying Time Limit}

Table~\ref{tab:varying_drone_time} reports the impact of increasing the drone flying time limit \(\tau\) on both makespan and computational runtime, with $\delta = 5$ and a fixed fleet of two trucks each carrying two drones. As \(\tau\) increases from 0.5 to 2.0 hours, drones can service more distant arcs and the makespan falls steadily in every instance. In the smallest network class (N50E100R15), for example, the objective decreases from 23.7 at \(\tau=0.5\) to 9.1 at \(\tau=2.0\), with a 61.6 \% reduction. However, the runtime increases from 0.3 minutes to 0.5 minutes. In the largest instance (N500E1000R100), extending \(\tau\) from 0.5 to 2.0 hours cuts the makespan by 57.9 \% but raises runtime by 71.3 \%.  

\begin{table}[]
\centering
\scriptsize
\caption{Impact of drone flying time on the solution and runtime}
\label{tab:varying_drone_time}
\begin{tabular}{crrrrrrrr}
\toprule
\multirow{2}{*}{Instance} & \multicolumn{2}{c}{$\tau = 0.5$}      & \multicolumn{2}{c}{$\tau = 1.0$}      & \multicolumn{2}{c}{$\tau = 1.5$}       & \multicolumn{2}{c}{$\tau = 2.0$}       \\
\cmidrule(lr){2-3} \cmidrule(lr){4-5} \cmidrule(lr){6-7} \cmidrule(lr){8-9}
         & \multicolumn{1}{c}{Obj} & \multicolumn{1}{c}{Time}  & \multicolumn{1}{c}{Obj} & \multicolumn{1}{c}{Time}  & \multicolumn{1}{c}{Obj} & \multicolumn{1}{c}{Time}  & \multicolumn{1}{c}{Obj} & \multicolumn{1}{c}{Time}  \\
\multicolumn{1}{l}{}      & \multicolumn{1}{c}{}    & \multicolumn{1}{c}{(min)} & \multicolumn{1}{c}{}    & \multicolumn{1}{c}{(min)} & \multicolumn{1}{c}{}    & \multicolumn{1}{c}{(min)} & \multicolumn{1}{c}{}    & \multicolumn{1}{c}{(min)} \\
\midrule
N50E100R15 & 23.7     & 0.3        & 20.5     & 0.4        & 13.8     & 0.5        & 9.1      & 0.5        \\
N50E100R20 & 29.8     & 0.6        & 26.8     & 0.8        & 18.7     & 0.9        & 12.1     & 1.0        \\
N100E200R25& 48.1     & 1.0        & 38.9     & 1.2        & 27.9     & 1.7        & 19.7     & 1.7        \\
N100E200R30& 52.8     & 1.7        & 45.1     & 2.3        & 32.5     & 2.6        & 21.6     & 2.8        \\
N200E400R40& 79.2     & 3.7        & 63.3     & 5.0        & 42.0     & 5.9        & 29.2     & 6.3        \\
N200E400R50& 91.5     & 6.7        & 78.0     & 9.4        & 52.8     & 10.9       & 36.3     & 11.5       \\
N300E600R50& 101.3    & 7.1        & 83.0     & 10.3       & 56.0     & 11.5       & 40.6     & 12.0       \\
N300E700R70& 129.8    & 17.6       & 108.0    & 25.7       & 74.7     & 28.1       & 54.5     & 30.3       \\
N400E800R80& 170.9    & 26.4       & 143.5    & 40.1       & 101.8    & 43.1       & 71.2     & 46.1       \\
N500E1000R100             & 206.3    & 53.4       & 174.5    & 83.2       & 122.1    & 85.9       & 86.9     & 91.5       \\
\bottomrule
\end{tabular}
\end{table}

On average across all ten instance classes, increasing \(\tau\) from 0.5 to 1.0 hour yields a 15.8 \% reduction in makespan at the cost of a 39.6 \% longer runtime. Similarly, increasing \(\tau\) further from 1.0 to 1.5 delivers a 30.8 \% drop in makespan with a 15.8 \% runtime penalty, and raising \(\tau\) from 1.5 to 2.0 brings another 30.7 \% improvement in routing efficiency for a 5.7 \% increase in runtime. Although today’s commercial delivery drones typically operate flights of around one hour, ongoing advances in drone battery capacity are rapidly extending their operational range, making higher $\tau$ values increasingly realistic in the near future.

\subsubsection{Analysis of $\delta$-hop Window}

\begin{figure}[!htbp]
    \centering
    \begin{subfigure}[b]{0.49\textwidth}
        \centering
        \includegraphics[scale=0.5]{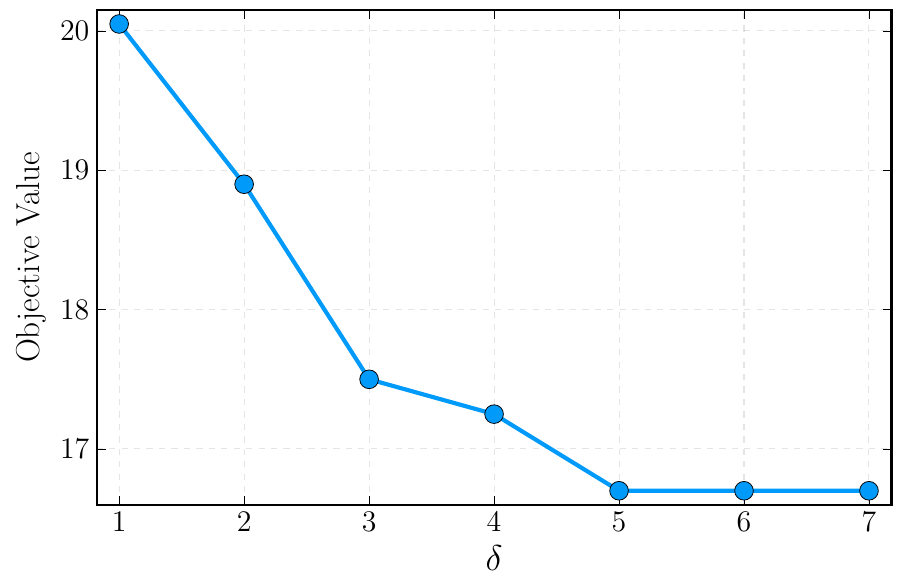}
        \caption{Instance with $N=50$, $E=100$, and $R=15$}
        \label{subfig:deltas_15}
    \end{subfigure}
    \hfill
    \begin{subfigure}[b]{0.49\textwidth}
        \centering
        \includegraphics[scale=0.5]{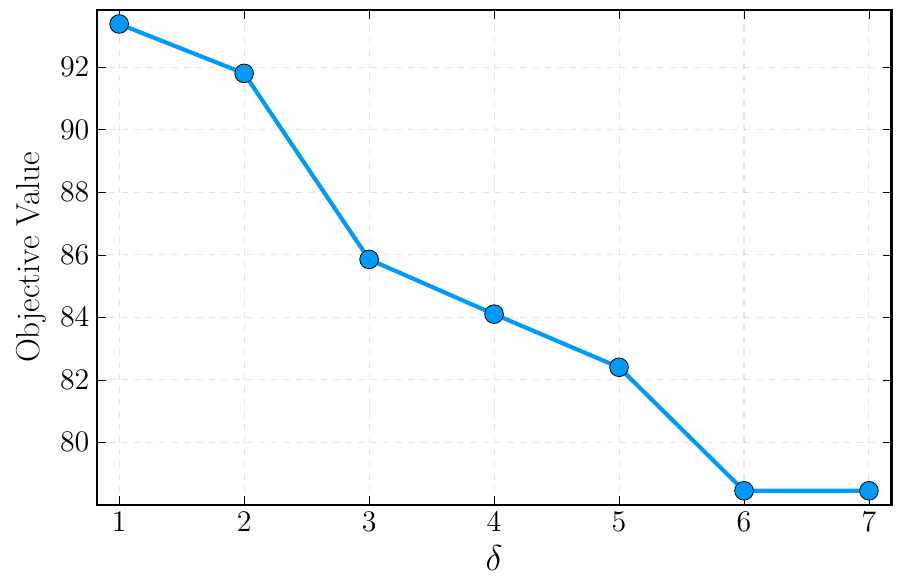}
        \caption{Instance with $N=200$, $E=400$, and $N=50$}
\label{subfig:deltas_50}
    \end{subfigure}
    \caption{Effect of varying $\delta$ on the makespan of two instances}
    \label{fig:varying_deltas}
\end{figure}

Using our proposed HGA, we analyze the effect of varying the parameter $\delta$. Figures~\ref{subfig:deltas_15} and~\ref{subfig:deltas_50} illustrate how adjusting the $\delta$-hop window influences the makespan for two instances belonging to classes N50E100R15 and N200E400N50, respectively. Recall that $\delta = 1$ implies each drone sortie departs from a truck stop and returns at the immediately subsequent truck stop, whereas higher $\delta$ values allow the truck to traverse up to $\delta$ arcs before rendezvousing with the drone. The experiments are conducted with a fixed fleet configuration $K = 2$ and $M = 2$, and drone flying time limit $\tau = 1.0$.

Generally, increasing $\delta$ reduces the makespan since drones can rendezvous further from their launch points. However, the incremental benefit diminishes as $\delta$ increases. Thus, practitioners should select an appropriate $\delta$ value by carefully balancing routing flexibility and makespan savings against operational complexity and synchronization constraints, as excessively large $\delta$ values may be practically infeasible.

\section{Conclusions} \label{sec:conclusions}

This paper introduced the Rural Postman Problem with multiple Trucks and Drones (RPP-mTD), a new arc-routing variant that captures multi-vehicle coordination in emerging truck–drone logistics.  
We developed a Hybrid Genetic Algorithm (HGA) with multiple neighborhood search methods to effectively solve this problem.  
Benchmark experiments on small‐scale instances with one truck and one drone show that HGA achieves an average optimality gap of 2.25\%, outperforming the state‐of‐the‐art ALNS (2.5\%) while delivering solutions significantly faster.
On larger networks, up to 500 nodes and 100 required arcs, the algorithm scales well and produces good quality solutions within minutes.

Our experiments yield three managerial insights. (1) Adding trucks consistently reduces makespan, whereas addition of drones per truck may not be useful for larger networks due to limited drone flying time. (2)  Extending drone flying time (\(\tau\)) sharply lowers makespan but can inflate runtime on very large instances. (3)  Relaxing the drone launch node constraint through a \(\delta\)-hop window further decreases the makespan; however, beyond moderate $\delta$ values, the incremental benefit diminishes.
Future research could explore extensions involving stochastic travel times and non-linear battery consumption. Another promising direction is to study scenarios where drones are not specifically associated with individual trucks, allowing them to fly in and out of different trucks throughout the operation.

\section*{Acknowledgments}

This work was supported by the National Science Foundation under award number 2032458 and the National Research Foundation of Korea grant (RS-2023-00259550) funded by the Ministry of Science and ICT. Part of this research was conducted during the first author’s doctoral program at the University of South Florida.

\bibliographystyle{ormsv080} 
\bibliography{ref}


\end{document}